\documentclass{article} 
\usepackage{iclr2026_conference,times}

\usepackage{amsmath,amsfonts,bm}

\def\eqref#1{equation~\ref{#1}}

\def\1{\bm{1}}

\DeclareMathAlphabet{\mathsfit}{\encodingdefault}{\sfdefault}{m}{sl}
\SetMathAlphabet{\mathsfit}{bold}{\encodingdefault}{\sfdefault}{bx}{n}

\usepackage{hyperref}
\usepackage{url}
\usepackage{booktabs}   
\usepackage{multirow}   
\usepackage{array}      
\usepackage{amsthm}

\usepackage{algorithm}      
\usepackage{algpseudocode}  
\usepackage{makecell}
\usepackage{pifont}
\usepackage{newtxtext}

\newcommand{\modelname}{\textsc{FantasyVLN}}

\usepackage[most]{tcolorbox}
\tcbset{
    mygraybox/.style={
        colback=gray!10,      
        colframe=gray!50,     
        arc=3mm,              
        boxrule=0.4pt,        
        left=6pt, right=6pt,  
        top=6pt, bottom=6pt,  
    }
}

\usepackage[utf8]{inputenc}

\usepackage{makecell}      
\usepackage[table]{xcolor} 
\usepackage[utf8]{inputenc}
\usepackage{amsmath}           
\usepackage{graphicx}          
\usepackage{booktabs}      
\usepackage{enumitem}
\usepackage{subcaption} 

\definecolor{lightgray}{gray}{0.9}
\newcommand{\B}[1]{\textbf{#1}}

\newcommand{\equal}{\textsuperscript{*}}           
\newcommand{\leader}{\textsuperscript{$\dagger$}}  
\newcommand{\corr}{\textsuperscript{$\ddagger$}}   
\newcommand{\intern}{\textsuperscript{$\S$}}       

\title{\modelname: Unified Multimodal Chain-of-Thought Reasoning for Vision-Language Navigation}

\author{
    Jing Zuo\textsuperscript{1,2}\equal\intern,\quad
    Lingzhou Mu\textsuperscript{1,3}\equal\intern,\quad
    Fan Jiang\textsuperscript{1}\equal\leader\corr,\quad
    Chengcheng Ma\textsuperscript{1},\\
    ~\B{Mu Xu}\textsuperscript{1},\quad
    \B{Yonggang Qi}\textsuperscript{2}\corr 
    \\
    \\
    \textsuperscript{1}Fantasy AIGC Team,\\
    \textsuperscript{2}Beijing University of Posts and Telecommunications,\\
    \textsuperscript{3}Tsinghua University \\
    \\
    {\footnotesize\texttt{jiangfan0576@gmail.com}}\\
    {\footnotesize\texttt{qiyg@bupt.edu.cn}}
}

\iclrfinalcopy 
\begin{document}

\maketitle
\begingroup
\renewcommand{\thefootnote}{\fnsymbol{footnote}}
\footnotetext[1]{Equal contribution.}
\footnotetext[2]{Project leader.}
\footnotetext[3]{Corresponding author.}
\footnotetext[4]{Work done during internship at Fantasy AIGC Team.}
\endgroup

\begin{abstract}
Achieving human-level performance in Vision-and-Language Navigation (VLN) requires an embodied agent to jointly understand multimodal instructions and visual-spatial context while reasoning over long action sequences. Recent works, such as NavCoT and NavGPT-2, demonstrate the potential of Chain-of-Thought (CoT) reasoning for improving interpretability and long-horizon planning. Moreover, multimodal extensions like OctoNav-R1 and CoT-VLA further validate CoT as a promising pathway toward human-like navigation reasoning. However, existing approaches face critical drawbacks: purely textual CoTs lack spatial grounding and easily overfit to sparse annotated reasoning steps, while multimodal CoTs incur severe token inflation by generating imagined visual observations, making real-time navigation impractical. In this work, we propose \modelname, a unified implicit reasoning framework that preserves the benefits of CoT reasoning without explicit token overhead. Specifically, imagined visual tokens are encoded into a compact latent space using a pretrained Visual AutoRegressor (VAR) during CoT reasoning training, and the model jointly learns from textual, visual, and multimodal CoT modes under a unified multi-CoT strategy. At inference, our model performs direct instruction-to-action mapping while still enjoying reasoning-aware representations. Extensive experiments on LH-VLN show that our approach achieves reasoning-aware yet real-time navigation, improving success rates and efficiency while reducing inference latency by an order of magnitude compared to explicit CoT methods. Project: \url{https://fantasy-amap.github.io/fantasy-vln/}
\end{abstract}

\begin{figure*}[!ht]
\centering
\includegraphics[width=1.\textwidth]{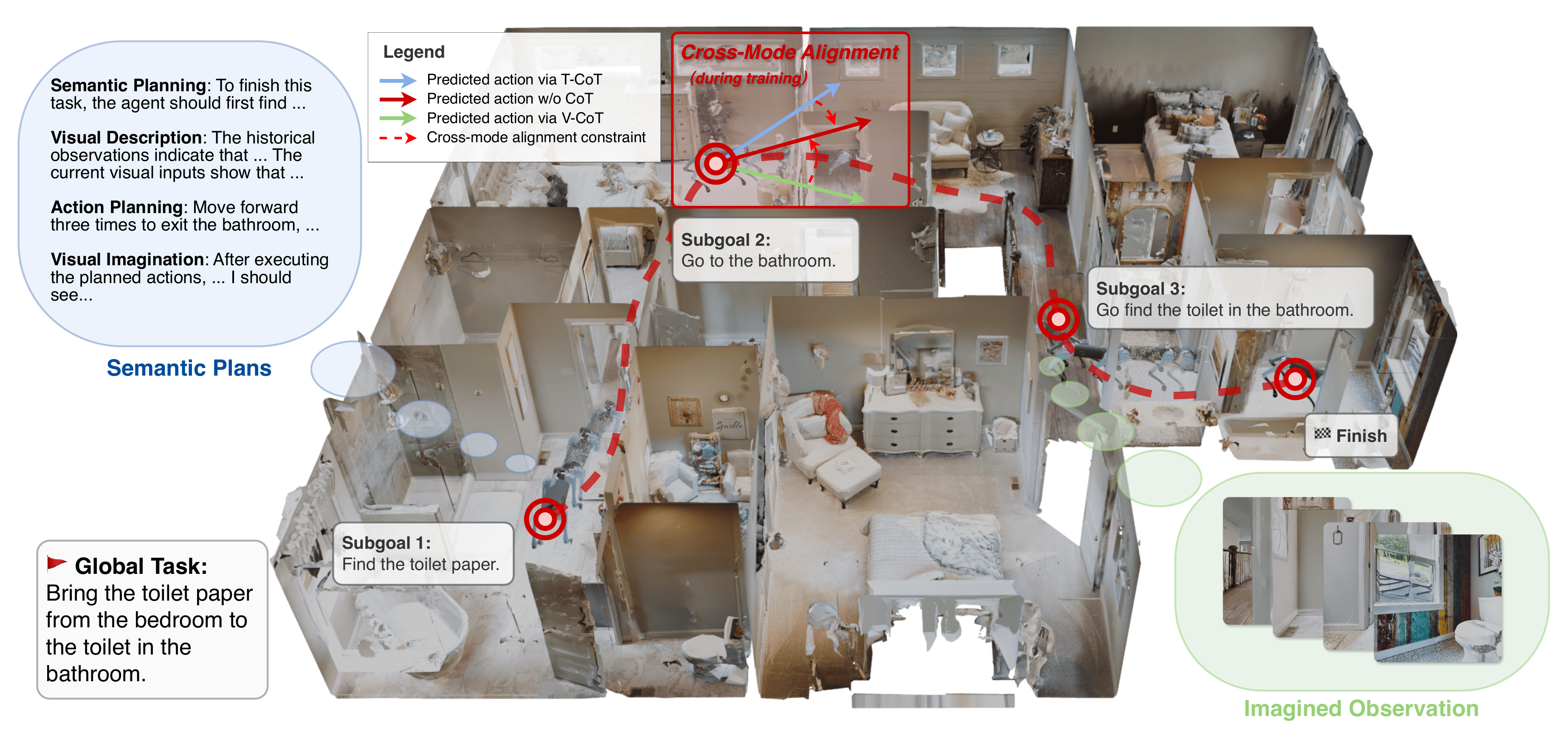}
\caption{Overview of \modelname. \modelname\ is a VLN framework that integrates the strengths of textual and visual CoT reasoning modes, thereby jointly modeling semantic planning and spatial understanding.}
\label{fig:overview}
\end{figure*}

\section{Introduction}
Vision-and-Language Navigation (VLN) aims to enable an embodied agent to follow natural-language instructions and navigate complex visual environments \cite{krantz2020beyond, wu2024vision, anderson2018vision, gu2022vision}. Solving this task requires the joint understanding of semantics from language and spatial geometry from visual observations, along with long-horizon reasoning to plan a sequence of actions. In particular, for multi-stage and long-horizon navigation scenarios as proposed in~\citep{song2025towards}, the ability to perform robust multimodal reasoning, i.e., to integrate linguistic intent with visual-spatial context over extended temporal dependencies, is especially critical. Despite the progress made by recent multimodal large models, achieving effective cross-modal reasoning in VLN remains challenging due to the semantic–spatial gap and the need for interpretable yet sample-efficient reasoning mechanisms.

The recent success of large language models (LLMs) has inspired the integration of Chain-of-Thought (CoT) reasoning into embodied navigation to improve interpretability and long-horizon decision-making. Methods such as NavCoT~\citep{lin2025navcot} and NavGPT-2~\citep{zhou2024navgpt} employ step-by-step textual reasoning to decompose navigation instructions or generate intermediate subgoals. However, their reasoning remains confined to the textual modality, typically by translating observations into captions, thereby limiting the joint modeling of semantic planning and spatial understanding, both essential for successful navigation. This limitation is compounded by the difficulty of annotating CoT supervision in VLN, as highlighted by EvolveNav~\citep{lin2025evolvenav}, where multiple valid action sequences often exist. Moreover, explicitly supervised CoT reasoning tends to overfit training distributions and generalize poorly to unseen environments.

Lately, works such as CoT-VLA~\citep{zhao2025cot}, VISTA~\citep{huang2025vista}, RBF++~\citep{chen2025rbf++}, OctoNav-R1~\citep{gao2025octonav}, and OmniNav~\citep{xue2025omninav} have extended CoT reasoning into visual or multimodal domains to better couple semantic and spatial reasoning for generalizability. While this multimodal CoT paradigm marks an important step forward, it also introduces new challenges for long-horizon navigation. In particular, modeling reasoning chains across both language and vision requires the model to iteratively generate and interpret imagined intermediate observations at each step, leading to severe token inflation. 
A typical reasoning step spanning 5--7 actions expands into over 3k--5k tokens, an order of magnitude larger than purely textual CoTs (usually $<$500 tokens).
This explosion in sequence length drastically increases both training and inference latency, rendering real-time navigation infeasible even on high-end GPUs.

To address these challenges, we propose a unified implicit reasoning framework that retains the benefits of CoT-style reasoning while eliminating its explicit token overhead during inference. The key idea is twofold: (i) During training, we encode the imagined observation tokens generated by multimodal CoT reasoning into a compact latent space using a pretrained Visual AutoRegressive (VAR) model. This significantly reduces sequence length and training cost without compromising the richness of visual reasoning. (ii) At inference, the agent performs direct instruction-to-action mapping while still leveraging reasoning-aware representations, inspired by the \emph{train-with-CoT, infer-without-CoT} paradigm of Aux-Think~\citep{wang2025auxthink}. 

Concretely, we introduce a unified multi-CoT training strategy that jointly learns from textual-only, visual-only, and textual–visual CoT modes using a special tag token to indicate each mode. This design unifies both the input format and model parameters within a single framework. During training, we align the action predictions from CoT-based reasoning modes with those from direct prediction (without CoT), enforcing modality-invariant reasoning representations. Consequently, the model learns implicit reasoning capabilities that generalize effectively without explicit CoT supervision or overfitting to training distributions.

To this end, our contributions are summarized as follows: (i) We propose the first unified implicit CoT reasoning framework that integrates textual, visual, and multimodal CoT paradigms within a single model. Unlike prior explicit CoT methods, our approach trains with diverse reasoning modes but performs inference without generating CoT sequences, achieving reasoning-aware yet real-time navigation. (ii) We introduce a gating-based multi-CoT learning mechanism that allows seamless switching among reasoning modes and direct action prediction under shared parameters. By aligning CoT-driven and direct action predictions, our model learns consistent, modality-invariant reasoning representations. (iii) To reduce the token explosion in multimodal reasoning, we compress imagined observation tokens into a compact latent space using a pretrained Visual AutoRegressor (VAR), improving training efficiency while preserving semantic–spatial reasoning capacity. (iv) Extensive experiments on the challenging LH-VLN benchmark demonstrate that our method substantially improves navigation success and efficiency in multi-stage and long-horizon scenarios, while reducing inference latency by an order of magnitude compared to explicit CoT approaches.

\begin{figure*}[!htbp]
\centering
\includegraphics[width=1.\textwidth]{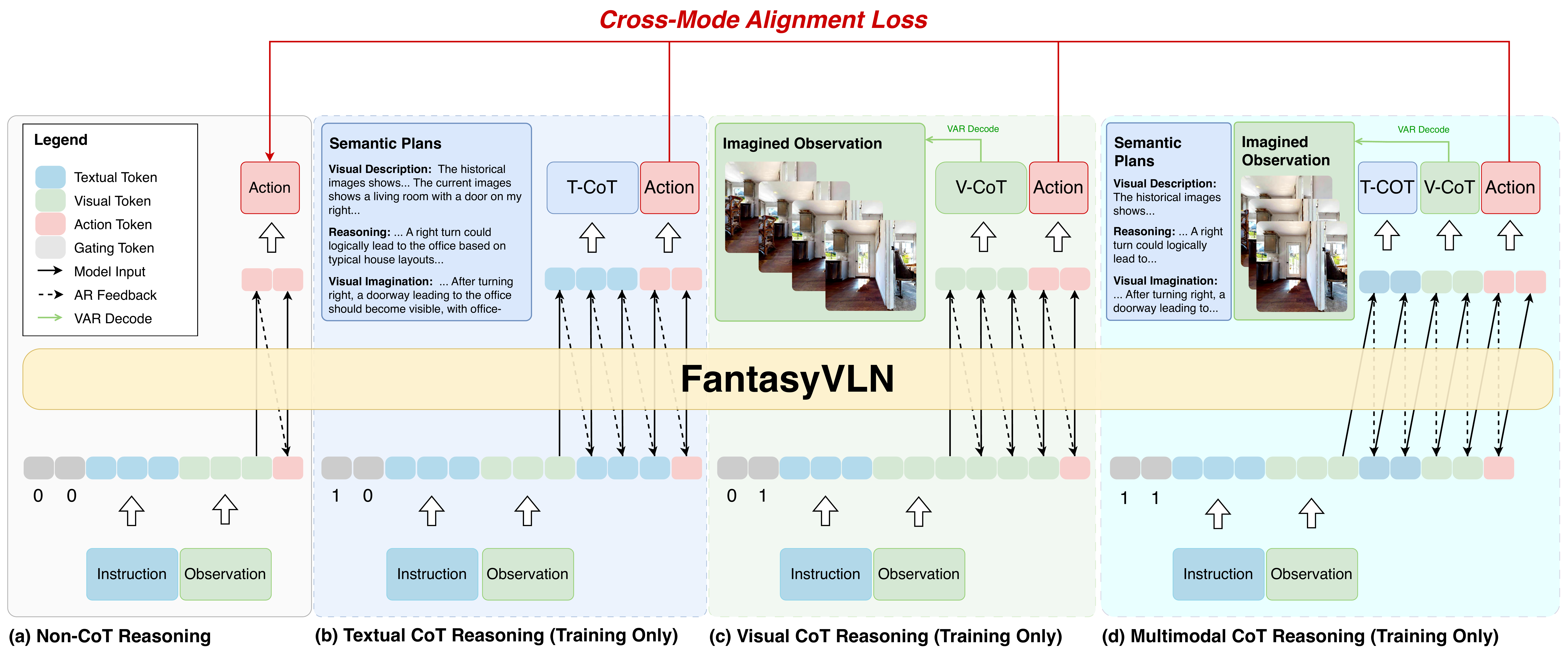}
\caption{Overview of our unified multimodal Chain-of-Thought reasoning framework. The model supports four reasoning modes under a shared architecture: (a) non-CoT reasoning for real-time inference, (b) textual CoT, (c) visual CoT enabled by VAR-compressed imagined observations, and (d) multimodal CoT combining textual and visual reasoning. A gating mechanism switches the model across reasoning modes, while the action predictions from CoT modes are consistently aligned with the non-CoT mode.}
\label{fig:framework}
\end{figure*}

\section{Related Works}
\label{sec:related-works}
\subsection{Vision-and-Language Navigation}
Early VLN models typically separate perception, instruction understanding, and action planning into discrete modules, and rely on imitation~\citep{nguyen2019vision, wang2022towards, wu2020towards} or reinforcement learning~\citep{xu2023benchmarking, wang2020vision} with auxiliary tasks such as progress monitoring or instruction reweighting.
However, these methods, built on panoramic observations in discrete environments (e.g., R2R~\citep{anderson2018vision} and RxR~\citep{ku2020room}), suffer from poor semantic alignment and limited generalization in continuous or unseen environments (e.g. VLN-CE~\citep{krantz2020beyond}).
To address these limitations, recent studies have shifted toward end-to-end navigation policy learning with pretrained vision-language models.
For example, Poliformer~\citep{zeng2025poliformer} introduces transformer-based on-policy reinforcement learning for video-level navigation.
NaVid~\citep{zhang2024navid} and Uni-NaVid~\citep{zhang2024uni} extend this paradigm by performing monocular video-based navigation without depth or maps and unifying multiple embodied tasks.
NaVILA~\citep{cheng2024navila} further integrates VLN with legged robot locomotion, achieving impressive cross-embodiment generalization.
While achieving remarkable progress on short-term tasks, they still struggle to reason and plan for long-horizon, multi-stage tasks.
More recently, CoT reasoning has emerged as a crucial paradigm for Embodied AI tasks.  
In VLN, NavGPT leverages the zero-shot CoT reasoning ability of GPT-4, while Aux-Think introduces auxiliary CoT supervision to internalize reasoning patterns during training.
Yet, existing CoT-based VLN methods confine reasoning within a single modality, leaving multimodal CoT largely unexplored.
In this paper, we follow the continuous environment setting and systematically investigate multimodal CoT reasoning in VLN.

\subsection{Chain-of-Thought Reasoning}
Chain-of-Thought (CoT) reasoning enables large language models (LLMs) to solve complex problems by explicitly generating intermediate steps~\citep{wei2022chain}.
Subsequent variants, such as Self-Consistency~\citep{wang23self} and Least-to-Most Prompting~\citep{zhou23least}, further enhance reasoning robustness and compositionality.
Recent studies have extended CoT to vision-language models~\citep{zhang24multimodal}, which can be categorized into three types based on the modality of reasoning steps: Textual CoT, Visual CoT, and Multimodal CoT.
Specifically, Textual CoT~\citep{zhang2024multimodal} in VLMs typically follows the format of vanilla LLM CoT.
Visual CoT methods, such as CoT-VLA~\citep{zhao2025cot} and DreamVLA~\citep{zhangdreamvla}, generate future frames before action prediction in manipulation tasks, while Multimodal CoT~\citep{cheng2025visual} jointly predicts paired textual and visual reasoning steps in multimodal tasks.
To the best of our knowledge, \modelname\ is the first unified CoT reasoning framework that integrates these three reasoning paradigms.

\section{Methods}
\label{sec:methods}
\subsection{Overview}
We propose \modelname, a VLN framework that integrates multimodal reasoning modes as its core design, while enabling implicit reasoning for efficient inference. As shown in Figure~\ref{fig:framework},
\modelname \ internalizes diverse CoT reasoning patterns across modalities through end-to-end joint training, and enhances the non-CoT reasoning mode via a cross-mode alignment constraint.
This enables combining the advantages of both textual and visual CoT reasoning without incurring explicit CoT reasoning time.
Moreover, we perform visual CoT reasoning in the latent space of the VAR model~\citep{tian2024visual}, which improves training and inference efficiency compared with pixel-space methods.

Below, we introduce the problem setup, cross-mode alignment constraint, unified multimodal implicit reasoning and latent visual CoT learning.

\subsection{Problem Setup}
VLN aims to develop an embodied agent \(\pi_\theta\) that navigates continuous 3D environments \(\mathcal{O}\) based on a natural language instruction \(\mathcal{I}\) and visual observations, which can be formulated as a non-Markovian temporal decision problem.
Let \(s_0\) denote the initial state, i.e., location and orientation, and \(\mathcal{U}\) denote the action space.
At each timestep \(t\), the agent \(\pi_\theta\) receives multi-view visual observations \(o_t \in \mathcal{O}\) and predicts future actions \(\mathcal{A}_t \in \mathcal{U}\) conditioned on the instruction \(\mathcal{I}\) and historical observations \(\{ o_{\le t} \}\).
Subsequently, the predicted actions \(\mathcal{A}_t\) are executed, transferring the agent \(\pi_\theta\) to a new state according to the environment dynamics.
This interaction process continues until a stop action is executed or the maximum step \(T\) is reached.

\subsection{Compact Visual Chain-of-Thought}
\label{sec:var}
Conventional V-CoT reasoning predicts thousands of visual tokens at each reasoning step, resulting in low training efficiency and high inference latency.
To address this issue, we present \textit{Compact Visual Chain-of-Thought} (CompV-CoT), which trains \textit{Qwen2.5-VL} to directly generate a compact set of visual tokens in the latent space of a pretrained VAR model, yielding a novel compressed visual chain-of-thought representation with far fewer tokens.
The VAR model follows a next-scale prediction paradigm to hierarchically encode visual information, achieving higher efficiency than conventional autoencoding approaches such as VAE, VQ-VAE~\citep{gafni2022make} or RAE~\citep{zheng2025diffusion}.
Given a \(256 \times 256\) image, the VAR model enables precise reconstruction using the corresponding low-scale representations, which contain only \(30\) visual tokens.
As shown in Table~\ref{table:compressors}, VAR achieves a higher compression ratio under comparable reconstruction quality.

\begin{table}[h]
\centering
\caption{Comparison of compression ratio and reconstruction error (MSE) across different visual compressors.}
\label{table:compressors}
\begin{tabular}{ccc}
\toprule
\textbf{Compressors} & \textbf{Comp. Ratio} & \textbf{MSE} \\
\midrule
RAE-DINOv2-B & 1/256 & 0.012 \\
RAE-SigLIP2-B & 1/256 & 0.011 \\
VAE & 1/64 & \textbf{0.005} \\
VQVAE & 1/64 & 0.007 \\
VAR & \textbf{1/2185} & 0.039 \\
\bottomrule
\end{tabular}
\end{table}

Specifically, we employ the VAR model as the visual decoder of the VLM and perform V-CoT in the VAR latent space.
The VLM first takes navigation instructions and visual observations as input and then generates latent representations of future observations before predicting actions.
The VAR model finally decodes generated representations into pixel frames.
We freeze the VAR model during training, while the VLM first learns to predict latent future observations and then infers the corresponding actions.
During inference, we use only the VLM to perform visual CoT-based navigation without explicit VAR decoding.
Owing to the highly efficient visual information compression and non-display image decoding, the proposed CompV-CoT method improves both training and inference efficiency.

\subsection{Unified Multimodal Chain-of-Thought}
\label{sec:um-cot}
Building on CompV-CoT, we further present a \textit{Unified Multimodal Chain-of-Thought} (UM-CoT) framework that integrates textual, compressed visual, and multimodal reasoning within a single agent.

\paragraph{Textual CoT in VLN.}
Textual CoT (T-CoT) models the agent’s reasoning as an explicit semantic planning process that bridges language understanding and action decision.
Instead of directly mapping instructions to actions, the agent first generates textual intermediate reasoning steps $\widehat{\mathcal{T}}_t$.
These reasoning steps then provide structured causal guidance for predicting subsequent actions $\widehat{\mathcal{A}}_t$, enabling interpretable and more reliable decision-making.
Specifically, the intermediate steps typically involve inferring subgoals from the instruction, assessing progress through current and historical visual observations, and identifying actionable cues for achieving the next objective.

\paragraph{CompV-CoT as Visual CoT.}
For visual CoT (V-CoT), we directly adopt the CompV-CoT introduced in Sec.~\ref{sec:var} as the visual reasoning mode in UM-CoT.
In this setting, the agent imagines future observations in the VAR latent space by predicting compressed visual tokens, and then infers actions conditioned on the imagined latent trajectory.
Compared with pixel-space prediction, this CompV-CoT design yields more efficient and stable visual reasoning.

\paragraph{Multimodal CoT in VLN.}
Multimodal CoT (MM-CoT) is defined as a native combination of T-CoT and CompV-CoT, where the agent is required to generate paired textual–visual reasoning steps.
We denote the multimodal reasoning trace as $\mathcal{M}_t = [\mathcal{T}_t, \mathcal{V}_t]$, which jointly encodes semantic plans and imagined future observations, and use it to guide subsequent action prediction.

\paragraph{Unified Multimodal CoT Reasoning Framework.}
To unify the above reasoning modes within a single framework, we introduce two binary gating signals $g_{\mathcal{T}}$ and $g_{\mathcal{V}}$ that control whether textual and visual reasoning is activated.
Given $\mathcal{I}$, $\{o_{\le t}\}$, and $(g_{\mathcal{T}}, g_{\mathcal{V}})$, the agent jointly predicts reasoning traces and actions:
\begin{equation}
[\widehat{\mathcal{R}}_t, \widehat{\mathcal{A}}_t] = \pi_\theta \big( \mathcal{I}, \{o_{\le t}\}, g_{\mathcal{T}}, g_{\mathcal{V}} \big),
\end{equation}
where
\begin{equation}
\label{eq:unified_modeling}
\widehat{\mathcal{R}}_t =
\begin{cases}
\text{None}, & \text{if } (g_{\mathcal{T}}, g_{\mathcal{V}}) = (0, 0), \\[4pt]
\widehat{\mathcal{T}}_t, & \text{if } (g_{\mathcal{T}}, g_{\mathcal{V}}) = (1, 0), \\[4pt]
\widehat{\mathcal{V}}_t, & \text{if } (g_{\mathcal{T}}, g_{\mathcal{V}}) = (0, 1), \\[4pt]
\widehat{\mathcal{M}}_t, & \text{if } (g_{\mathcal{T}}, g_{\mathcal{V}}) = (1, 1).
\end{cases}
\end{equation}
This gating mechanism allows a single policy to flexibly operate in 
CoT, T-CoT, CompV-CoT, and MM-CoT modes.

\paragraph{Joint Training via Data Mixture.}
Given the navigation instruction $\mathcal{I}$, visual observations $\{o_{\le t}\}$ and the ground truth action $\mathcal{A}_t$.
To enable end-to-end training, we organize the expert navigation dataset $\mathcal{D}$ into five-tuples:
\begin{equation}
[\mathcal{I}, \{o_{\le t}\}, \mathcal{T}_t, \mathcal{V}_t, \mathcal{A}_t] \in \mathcal{D},
\end{equation}
where $\mathcal{T}_t$ and $\mathcal{V}_t$ denote the ground truth textual reasoning steps and CompV-CoT visual reasoning steps, respectively.
We employ \textit{Qwen-VL-Max} to generate textual reasoning traces $\mathcal{T}_t$.
During training, $(g_{\mathcal{T}}, g_{\mathcal{V}})$ are uniformly sampled and integrated with $\mathcal{I}$ and $\{o_{\le t}\}$ to form the query, while the answer is constructed according to Eq.~(\ref{eq:unified_modeling}) by selecting $\mathcal{R}_t \in \{\text{None}, \mathcal{T}_t, \mathcal{V}_t, \mathcal{M}_t\}$ together with $\mathcal{A}_t$.
The joint objective is defined as:
\begin{equation}
\label{eq:joint_training}
\begin{aligned}
\mathcal{L}_{\text{Joint}} 
& = (\neg g_{\mathcal{T}} \land \neg g_{\mathcal{V}})\, \mathcal{L}_{\text{CE}} \big( \widehat{\mathcal{A}}_t, \mathcal{A}_t \big) \\
& + (g_{\mathcal{T}} \land \neg g_{\mathcal{V}})\, \mathcal{L}_{\text{CE}} \big( [ \widehat{\mathcal{T}}_t, \widehat{\mathcal{A}}_t ], [ \mathcal{T}_t, \mathcal{A}_t ] \big) \\
& + (\neg g_{\mathcal{T}} \land g_{\mathcal{V}})\, \mathcal{L}_{\text{CE}} \big( [ \widehat{\mathcal{V}}_t, \widehat{\mathcal{A}}_t ], [ \mathcal{V}_t, \mathcal{A}_t ] \big) \\
& + (g_{\mathcal{T}} \land g_{\mathcal{V}})\, \mathcal{L}_{\text{CE}} \big( [ \widehat{\mathcal{M}}_t, \widehat{\mathcal{A}}_t ], [ \mathcal{M}_t, \mathcal{A}_t ] \big),
\end{aligned}
\end{equation}
where $\mathcal{L}_{\text{CE}}$ denotes the causal cross-entropy loss.

\begin{algorithm}[htbp]
\caption{Cross-Mode Aligned Joint Training}
\label{alg:cross-mode_aligned_joint_training}
\begin{algorithmic}[1]
\State \textbf{Input:} Dataset $\mathcal{D}$, parameters $\theta$, learning rate $\eta$, alignment weight $\lambda_{\text{align}}$
\State \textbf{Output:} Trained parameters $\theta^*$
\While{not converged}
  \State $[\mathcal{I}, \{o_{\le t}\}, \mathcal{T}_t, \mathcal{V}_t, \mathcal{A}_t] \sim \mathcal{D}$
  \State $\widehat{\mathcal{A}}_t \leftarrow \pi_\theta(\mathcal{I}, \{o_{\le t}\}, g_{\mathcal{T}}{=}0, g_{\mathcal{V}}{=}0)$
  \State $\theta \leftarrow \theta - \eta \nabla_\theta \mathcal{L}_{\text{CE}}(\widehat{\mathcal{A}}_t, \mathcal{A}_t)$
  \State $\widetilde{\mathcal{A}}_t \leftarrow \text{sg} \big[\pi_\theta(\mathcal{I}, \{o_{\le t}\}, g_{\mathcal{T}}{=}0, g_{\mathcal{V}}{=}0)\big]$
  \State $[\widehat{\mathcal{T}}_t, \widehat{\mathcal{A}}_t^{\mathcal{T}}] \leftarrow \pi_\theta(\mathcal{I}, \{o_{\le t}\}, g_{\mathcal{T}}{=}1, g_{\mathcal{V}}{=}0)$
  \State $[\widehat{\mathcal{V}}_t, \widehat{\mathcal{A}}_t^{\mathcal{V}}] \leftarrow \pi_\theta(\mathcal{I}, \{o_{\le t}\}, g_{\mathcal{T}}{=}0, g_{\mathcal{V}}{=}1)$
  \State $[\widehat{\mathcal{M}}_t, \widehat{\mathcal{A}}_t^{\mathcal{M}}] \leftarrow \pi_\theta(\mathcal{I}, \{o_{\le t}\}, g_{\mathcal{T}}{=}1, g_{\mathcal{V}}{=}1)$
  \State \text{Compute} \(\mathcal{L}_{\text{Joint}}^*\) using Eq.~(\ref{eq:aligned_joint_opt})
  \State $\theta \leftarrow \theta - \eta \nabla_\theta \mathcal{L}_{\text{Joint}}^*$
\EndWhile
\State $\theta^* \leftarrow \theta$
\State \Return $\theta^*$
\end{algorithmic}
\end{algorithm}

\subsection{Cross-Mode Alignment Constraint}
To prevant conflict between different reasoning modes, we introduce a \textit{Cross-Mode Alignment Constraint} that regularizes the unified multimodal CoT training.
The key idea is to use the non-CoT reasoning mode as a supervisory signal to align all CoT variants, thereby embedding diverse reasoning behaviors into a shared latent policy.
Let \(\widehat{\mathcal{A}}_t, \widehat{\mathcal{A}}_t^{\mathcal{T}}, \widehat{\mathcal{A}}_t^{\mathcal{V}}\), and \(\widehat{\mathcal{A}}_t^{\mathcal{M}}\) denote the action predictions from the non-CoT, T-CoT, V-CoT, and MM-CoT reasoning modes, respectively.
In each iteration, we first optimize the non-CoT reasoning mode with the objective:
\begin{equation}
\label{eq:noncot_opt}
\mathcal{L}_{\text{non-CoT}} = \mathcal{L}_{\text{CE}} \big(\widehat{\mathcal{A}}_t, \mathcal{A}_t \big),
\end{equation}
where
\begin{equation}
\label{eq:noncot_forward}
\widehat{\mathcal{A}}_t = \pi_\theta \big( \mathcal{I}, \{o_{\le t}\}, g_{\mathcal{T}} = 0, g_{\mathcal{V}} = 0 \big).
\end{equation}
We then obtain the soft targets \(\widetilde{\mathcal{A}}_t\) by rerunning the forward process~(\ref{eq:noncot_forward}).
Finally, we incorporate the cross-mode alignment constraint into the joint objective of unified multimodal CoT reasoning:
\begin{equation}
\label{eq:aligned_joint_opt}
\mathcal{L}_{\text{Joint}}^* = \mathcal{L}_{\text{Align}} + \mathcal{L}_{\text{CoT}},
\end{equation}
where 
\begin{equation}
\begin{aligned}
& \mathcal{L}_{\text{Align}} = \mathcal{L}_{\text{CE}} \big( \widehat{\mathcal{A}}_t^{\mathcal{T}}, \widetilde{\mathcal{A}}_t \big) \\
& \quad + \mathcal{L}_{\text{CE}} \big( \widehat{\mathcal{A}}_t^{\mathcal{V}}, \widetilde{\mathcal{A}}_t \big) + \mathcal{L}_{\text{CE}} \big( \widehat{\mathcal{A}}_t^{\mathcal{M}}, \widetilde{\mathcal{A}}_t \big),
\end{aligned}
\end{equation}
and
\begin{equation}
\begin{aligned}
& \mathcal{L}_{\text{CoT}} = \mathcal{L}_{\text{CE}} \big( \widehat{\mathcal{T}}_t, \mathcal{T}_t \big) \\
& \quad + \mathcal{L}_{\text{CE}} \big( \widehat{\mathcal{V}}_t, \mathcal{V}_t \big) + \mathcal{L}_{\text{CE}} \big( \widehat{\mathcal{M}}_t, \mathcal{M}_t \big).
\end{aligned}
\end{equation}
We alternately minimize the non-CoT objective~(\ref{eq:noncot_opt}) and the cross-mode aligned joint objective~(\ref{eq:aligned_joint_opt}) until the training losses \(\mathcal{L}_{\text{non-CoT}}\) and \(\mathcal{L}_{\text{Joint}}^*\) converge.
During this alternating optimization, all reasoning modes operate on similar inputs, share network parameters, and are aligned to identical supervisory signals, thereby implicitly embedding diverse CoT reasoning patterns into a unified latent representation.
The overall algorithm is presented in Algorithm~\ref{alg:cross-mode_aligned_joint_training}.

\subsection{VLN During Inference}
Due to the real-time demands of VLN and the inference latency introduced by explicit CoT token decoding, we follow Aux-Think~\citep{wang2025auxthink} and adopt the non-CoT reasoning mode during inference.
Similar to Aux-Think~\citep{wang2025auxthink}, our framework serves as an implicit reasoning mechanism that internalizes diverse CoT patterns and implicitly enhances non-CoT reasoning through cross-mode alignment and joint training across reasoning modes.


\begin{table}[tbp]
\centering
\small
\setlength{\tabcolsep}{6pt}
\renewcommand{\arraystretch}{1.2}
\caption{Comparison of navigation accuracy across different VLN methods on LH-VLN benchmark.  The best and second-best results are marked in bold and underlined, respectively.}
\resizebox{0.475\textwidth}{!}{
\begin{tabular}{cccccc}
\toprule
 \textbf{CoT Modal} & \textbf{Methods} & \textbf{SR} & \textbf{ISR} & \textbf{CSR} & \textbf{CGT} \\
\midrule
\multirow{5}{*}{None/ZS} 
 & Random & 0 & 0 & 0 & 0 \\
 & GLM-4v prompt & 0 & 0 & 0 & 0 \\
 & GPT-4 + NaviLLM & 0 & 2.19 & 1.45 & 2.61 \\
 & MGDM & 0 & 2.34 & 1.65 & \underline{2.91} \\
\midrule
\multirow{2}{*}{Visual}
 & CoT-VLA & 0 & 0 & 0 & 0 \\
 & WorldVLA & 0 & 0 & 0 & 0 \\
\midrule
\multirow{1}{*}{Textual}
 & Aux-Think & \underline{0.65} & \underline{3.16} & \underline{2.04} & 1.47 \\
\midrule
\multirow{1}{*}{unified multimodal}
 & \modelname \ & \textbf{2.44} & \textbf{11.01} & \textbf{9.64} & \textbf{8.99} \\
\bottomrule
\end{tabular}
}
\label{tab:main_res}
\end{table}

\begin{table}[!b]
\centering
\small
\setlength{\tabcolsep}{6pt}
\renewcommand{\arraystretch}{1.2}
\caption{Comparison of navigation accuracy across different reasoning mode combinations on LH-VLN.}
{
\begin{tabular}{cccc|cccc} 
\toprule
\textbf{non-CoT} & \textbf{T-CoT} & \textbf{V-CoT} & \textbf{MM-CoT} & \textbf{SR} & \textbf{ISR} & \textbf{CSR} & \textbf{CGT} \\
\midrule
\checkmark & & & & 0 & 2.01 & 1.51 & 1.55 \\
\checkmark & \checkmark & & & 0.98 & 8.26 & 6.60 & 6.15 \\
\checkmark &  & \checkmark & & 1.46 & \textbf{11.19} & \textbf{9.66} & 8.84 \\
\checkmark &  &  & \checkmark & 0.49 & 7.77 & 6.48 & \underline{8.89} \\
\checkmark & \checkmark & \checkmark & \checkmark & \textbf{2.44} & 11.01 & 9.64 & \textbf{8.99} \\
\bottomrule
\end{tabular}
}
\label{tab:mode_combination}
\end{table}

\section{Experiments}
\label{sec:experiments}
\subsection{Experimental Setup}
\paragraph{Benchmark.}
We evaluate \modelname \ on the challenging LH-VLN~\citep{song2025towards} benchmark, which is characterized by multi-stage tasks and long navigation trajectories.
On one hand, multi-stage navigation requires the agent to sequentially reach multiple goals, imposing higher demands on reasoning and planning.
On the other hand, longer navigation trajectories amplify cumulative errors compared to shorter ones.
Following the standard LH-VLN setting, we perform online evaluation on the test set, where both the tasks and scenes are unseen.

\paragraph{Baselines.}
We compared the proposed \modelname \ with several representative methods. They can be divided into four categories:
(i) textual CoT-based methods, Aux-Think~\citep{wang2025auxthink};
(ii) visual CoT-based methods, CoT-VLA and WorldVLA;
(iii) memory-based methods, MGDM;
(iv) other baselines provided by LH-VLN, GLM-4v prompt, NaviLLM and GPT-4 + NaviLLM.
For fair comparison, all methods are trained on the same LH-VLN training set, and the validation set is used to select the best checkpoint for each method.
For Aux-Think and CoT-VLA, we implement their methods based on the descriptions in their papers, as the training codes are not publicly available.
For WorldVLA, we adapt the official implementation by modifying the preprocessing pipeline to support training on the LH-VLN dataset.
For all other methods, we use the implementations provided by LH-VLN.

\paragraph{Metrics.}
Following~\citep{song2025towards}, we use Success Rate (SR), Independent Success Rate (ISR), Conditional Success Rate (CSR), and CSR weighted by Ground Truth (CGT) to measure multi-stage navigation accuracy.
SR denotes the success rate of multi-stage task navigation, ISR represents the success rate of individual subtasks, CSR weights the ISR according to the success of preceding subtasks, and CGT further weights the CSR based on the length of the expert trajectory.
Moreover, we introduce the Action Per Second (APS) to evaluate inference efficiency:
\begin{equation}
\text{APS} = \frac{N_\text{act}}{T_\text{nav}},
\end{equation}
where $N_\text{act}$ denotes the total number of executed actions, and $T_\text{nav}$ represents the total navigation time in seconds.

\begin{table}[!t]
\centering
\caption{Comparison of inference efficiency across different CoT reasoning methods. The best results are marked in bold.}
{
\begin{tabular}{cccc}
\toprule
\textbf{Reasoning Mode} & \textbf{Methods} & \textbf{Model Size} & \textbf{APS} \\
\midrule
\multirow{1}{*}{Explicit}
 & CoT-VLA & 7B & 0.19 \\
\midrule
\multirow{3}{*}{Implicit}
 & WorldVLA & 7B & 1.02 \\
 & Aux-Think & 8B & 0.97 \\
 & \modelname \ & 7B & \textbf{1.03} \\
\bottomrule
\end{tabular}
}
\label{tab:efficiency}
\end{table}

\begin{figure}[!b]
\centering
\includegraphics[width=\linewidth]{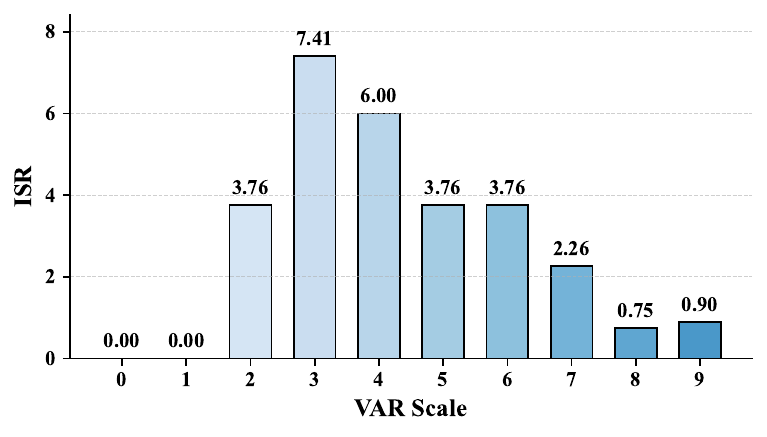}
\caption{ISR variation with respect to different VAR scales.}
\label{fig:isr_var_scale}
\end{figure}

\begin{figure*}[htbp ]
\centering
\includegraphics[width=\linewidth]{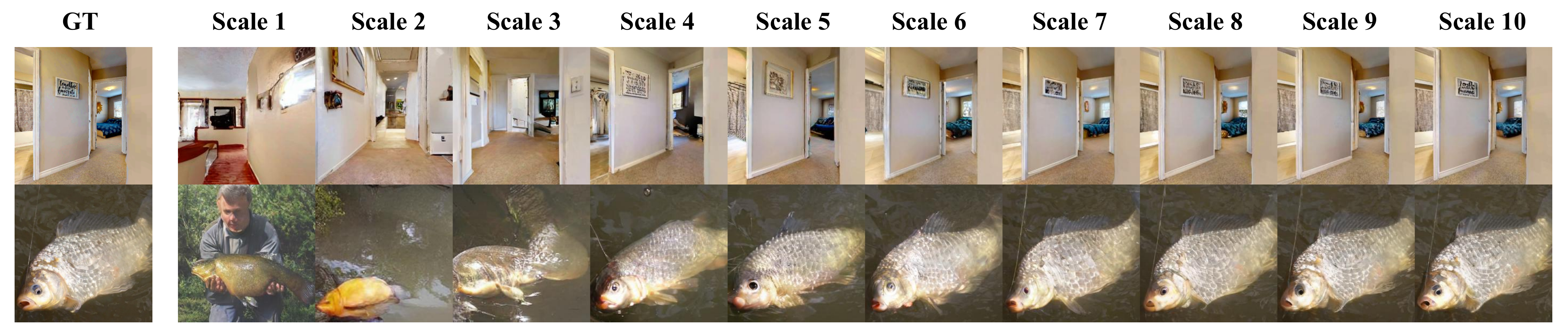}
\caption{Qualitative comparison of image reconstruction results produced by the VAR model using latent inputs across different scales. For each image, the VAR model receives the ground truth latents up to a specified scale and predicts all remaining scales; the final reconstruction is obtained by decoding the combined ground truth and predicted latents.}
\label{fig:var_rec}
\end{figure*}

\subsection{Main Results}
\paragraph{Navigation Accuracy.}
Table~\ref{tab:main_res} presents the quantitative results of navigation accuracy across different VLN methods on the LH-VLN benchmark.
\modelname \ achieves superior performance across all metrics, with SR, ISR, CSR, and CGT of 2.44, 11.01, 9.64, and 8.99, respectively, significantly surpassing all baselines.
Aux-Think shows suboptimal results in SR, ISR, and CSR, indicating that T-CoT enhances navigation robustness compared to non-CoT approaches.
However, its performance still exhibits a notable gap compared to \modelname, owing to the limitations of single-modal CoT modeling and the lack of an explicit–implicit alignment mechanism.
MGDM performs relatively well among non-CoT baselines, particularly in CGT, suggesting that memory mechanisms offer limited yet tangible benefits.
Overall, the results demonstrate that our unified multimodal implicit reasoning framework is crucial for tackling the complex multi-stage VLN task.

\paragraph{Inference Efficiency.}
To quantify the inference efficiency of different CoT reasoning methods, we report APS in Table~\ref{tab:efficiency}.
Implicit reasoning models, including \modelname, Aux-Think, and WorldVLA, exhibit comparable efficiency and outperform the explicit approach CoT-VLA by a substantial margin.
This outcome is expected.
Implicit reasoning predicts each action by decoding a single token, while explicit reasoning requires generating CoT reasoning steps with thousands of tokens.
Under similar model sizes, implicit CoT reasoning predicts approximately one action per second, while explicit CoT reasoning yields only 0.19 actions per second.
Therefore, implicit reasoning better satisfies the real-time requirements of the VLN task.

\subsection{Ablation Studies}
\paragraph{Contribution of Each Reasoning Mode.}
\modelname \ integrates diverse reasoning modes within a unified framework.
To verify the contribution of each reasoning mode to the overall framework, we explore various combinations of non-CoT, T-CoT, V-CoT, and MM-CoT modes during training.
As shown in Table~\ref{tab:mode_combination}, combining any CoT reasoning mode with non-CoT reasoning consistently improves navigation performance across all metrics.
Integrating all four reasoning modes further enhances the overall performance.

\paragraph{VAR Scale Selection.}
As detailed in Section~\ref{sec:var}, we perform V-CoT in the latent space of VAR.
To select the optimal VAR scale for latent V-CoT learning, we conduct comprehensive ablation studies on a subset of LH-VLN.
We first report the ISR results across different VAR scales, ranging from 1 to 10, as shown in Figure~\ref{fig:isr_var_scale}.
The results show that scale~4 achieves the best performance.
We attribute this to smaller scales lacking sufficient visual information, while larger scales leading redundancy.
To validate this argument, we randomly sample 100 images from LH-VLN and employ a pretrained VAR model to reconstruct them.
Specifically, VAR takes the ground truth latents up to a given scale as input and predicts the remaining latent scales.
The reconstructed images are then obtained by decoding both the ground truth and predicted latents together.
As shown in Figure~\ref{fig:var_rec}, the results are consistent with our argument.

\paragraph{Effect of Cross-Mode Alignment Constraint.}
We introduce a cross-mode alignment constraint into the joint training of \modelname.
Table~\ref{tab:cross-mode_alignment} compares SR, ISR, CSR, and CGT performance with and without this constraint.
Without the cross-mode alignment constraint (\ding{55}), \modelname \ exhibits weak navigation ability, achieving only marginal ISR, CSR, and CGT scores.
Adopting this constraint (\checkmark) yields substantial improvements across all metrics, with SR increasing from 0 to 2.44, ISR from 2.39 to 11.01, CSR from 1.19 to 9.64, and CGT from 1.28 to 8.99.
This indicates that cross-mode alignment is essential for \modelname, a unified framework that integrates diverse reasoning modes.

\begin{table}[htbp]
\centering
\caption{Comparison of SR, ISR, CSR, and CGT performance with and without cross-mode alignment.}
{
\begin{tabular}{c|cccc}
\toprule
\textbf{Alignment Constraint} & \textbf{SR} & \textbf{ISR} & \textbf{CSR} & \textbf{CGT} \\ 
\midrule
\ding{55}  & 0 & 2.39 & 1.19 & 1.28 \\
\checkmark & \textbf{2.44} & \textbf{11.01} & \textbf{9.64} & \textbf{8.99} \\
\bottomrule
\end{tabular}
}
\label{tab:cross-mode_alignment}
\end{table}

\subsection{More Results}
\begin{figure}[!ht]
\centering
\includegraphics[width=\linewidth]{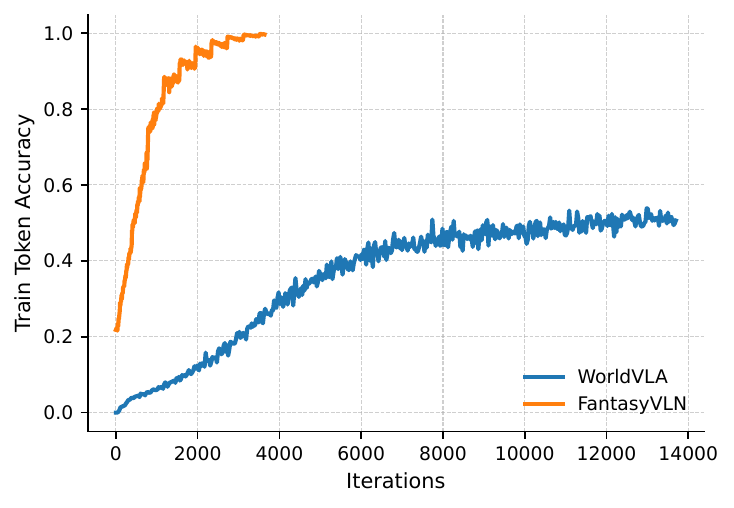}
\caption{Comparison of training efficiency between \modelname \ and WorldVLA.}
\label{fig:training_efficiency}
\end{figure}

\paragraph{Training Efficiency.}
As shown in Table~\ref{tab:main_res}, existing visual CoT methods (e.g., WorldVLA) achieve limited navigation accuracy on the LH-VLN benchmark, failing to generalize to long-horizon scenarios. To understand the underlying cause, we compare their training efficiency with our unified multimodal formulation. 
As shown in Figure~\ref{fig:training_efficiency}, WorldVLA exhibits slow and unstable convergence, requiring over $10$k iterations to reach moderate token prediction accuracy. This indicates that pixel-level V-CoT learning delivers weak gradient signals, as the model must reconstruct high-dimensional visual tokens for each reasoning step. In contrast, \modelname\ converges rapidly within a few thousand iterations, reflecting stable supervision and more efficient learning dynamics. This improvement stems from our CompV-CoT design, where visual reasoning operates in a compact latent space encoded by the pretrained VAR compressor. By replacing dense pixel reconstruction with compressed latent prediction, the model learns richer multimodal reasoning cues under a substantially lighter optimization burden.
Overall, these results highlight that CompV-CoT not only enhances reasoning efficiency but also yields more stable and interpretable learning behavior, contributing to the superior navigation accuracy of \modelname\ in long-horizon tasks.

\paragraph{Explicit vs. Implicit Reasoning.}
\modelname\ supports both explicit and implicit reasoning, enabling us to systematically compare their effectiveness across different CoT modalities. As summarized in Table~\ref{tab:implicit_vs_explicit}, we evaluate the T-CoT, V-CoT, and MM-CoT models under two inference modes. For each case, the model is jointly trained with its corresponding CoT mode and the direct prediction pathway. During inference, using the CoT branch corresponds to explicit reasoning, while employing the direct pathway corresponds to implicit reasoning.
Overall, implicit reasoning consistently yields higher navigation accuracy, particularly under multimodal settings. In MM-CoT, implicit inference achieves the best performance with $2.44$ SR and $11.01$ ISR, surpassing the explicit counterpart by a large margin. This result aligns with the observations in Aux-Think~\citep{wang2025auxthink}, suggesting that explicit CoT decoding may amplify cumulative reasoning errors across long trajectories. We attribute this phenomenon to two key factors: (i) the limited training data of LH-VLN (only $18$k trajectory slices of five steps each) makes explicit CoT sequences prone to overfitting and error propagation; (ii) explicit reasoning expands temporal dependencies, causing misaligned textual or visual CoT tokens to accumulate deviations over time. In contrast, implicit reasoning benefits from cross-mode alignment during training, allowing the model to internalize reasoning cues while maintaining stable and efficient inference.

\begin{table}[htbp]
\centering
\caption{Comparison of explicit and implicit CoT reasoning across modalities.}
\label{tab:implicit_vs_explicit}
\begin{tabular}{c|c|c|c|c}
\toprule
Metrics & Mode & T-CoT & V-CoT & MM-CoT \\
\midrule
\multirow{2}{*}{SR} 
& explicit & 0.98 & 0.49 & 0.98 \\
& implicit & 0.49 & 1.46 & 2.44 \\
\midrule
\multirow{2}{*}{ISR} 
& explicit & 8.26 & 7.34 & 8.62 \\
& implicit & 6.06 & 11.19 & 11.01 \\
\bottomrule
\end{tabular}
\end{table}

\section{Conclusion}
\label{sec:conclusion}
We introduced \modelname, a unified implicit reasoning framework that preserves the benefits of Chain-of-Thought supervision while avoiding the token explosion inherent to explicit textual or multimodal CoTs. By compressing imagined visual observations into a compact latent space via a pretrained VAR model and jointly training across textual, visual, and multimodal CoT modes under a unified multi-CoT strategy, the framework learns modality-invariant reasoning representations without requiring explicit CoT generation at inference. As a result, the agent performs direct instruction-to-action mapping while retaining reasoning-aware behavior. Experiments on the challenging LH-VLN benchmark show that this formulation substantially improves navigation accuracy and efficiency, while reducing inference latency by an order of magnitude compared to explicit CoT baselines. These findings demonstrate that implicit multimodal reasoning provides a practical pathway toward real-time embodied navigation, and highlight the potential of compact latent reasoning signals for closing the gap between semantic intent and spatial decision-making in complex environments.

{
    \small
    \bibliographystyle{iclr2026_conference}
    \bibliography{iclr2026_conference}
}


\clearpage
\setcounter{page}{1}

\setcounter{section}{0}

\appendix

\begin{center}
    \Large \textbf{Appendix}
\end{center}

\section{Data Preparation}

\subsection{Preprocessing}
An expert trajectory in LH-VLN comprises a temporal sequence of visual–action pairs \(\{o_t, a_t\}_{1:T}\) together with the natural-language task instruction \(\mathcal{I}\).
Since VLN is an online sequential decision-making problem with continual interaction with the environment, the agent \(\pi_\theta\) must act in real time based on its current and historical visual observations \(\{o_{\le t}\}\).
To construct training samples, we partition each navigation trajectory \(\mathcal{J}_i\) into non-overlapping slices.
Each slice contains the task instruction \(\mathcal{I}\), visual observations \(\{o_{\le t}\}\), and the future \(k\) actions \(\mathcal{A}_t\):
\begin{equation}
\label{eq:slice}
\{\mathcal{I},\{o_{\le t}\},\mathcal{A}_t\}_{t\in S_i}
\sim \text{Slice}(\mathcal{J}_i),\quad i=1,\ldots,N,
\end{equation}
where \(\mathcal{A}_t=\{a_t,a_{t+1},\ldots,a_{t+k-1}\}\), \(S_i=\{1,1+k,\ldots,T_i\}\), \(T_i\) is the number of actions in \(\mathcal{J}_i\), and \(N\) is the number of navigation trajectories in LH-VLN.
In practice, we set \(k = 5\).

\subsection{Navigation Prompt}
We use the following prompts to enable instruction-driven navigation behaviors across different reasoning modes.
For single-stage task, we set the prompt as:
\begin{tcolorbox}[mygraybox]
You are an autonomous navigation robot. You will get a task with historical and current pictures you see. Based on this information, you need to decide your next 5 actions, which could involve \texttt{\texttt{<|left|>}}, \texttt{<|right|>},\texttt{<|forward|>}. If you finish your mission, output \texttt{<|stop|>}. Your current observations are left side: \texttt{<image>}, front side: \texttt{<image>}, right side: \texttt{<image>}. Your historical pictures are: \texttt{<image>} ... \texttt{<image>}. Your mission is: [instruction].
\end{tcolorbox}

In multi-stage navigation tasks, the agent must stop upon completing each subtask and maintain awareness of how many subtasks have been finished. To this end, we further extend the prompt with the following description:
\begin{tcolorbox}[mygraybox]
PS: The mission is complex. You may infer several subtasks within the mission, and output \texttt{<|stop|>} when a sub-task is achieved. So far, you have output \texttt{<|stop|>} 0 times. Historical information reflects progress up to the current subgoal.
\end{tcolorbox}

\subsection{T-CoT Data Annotation}
We employ \texttt{Qwen-VL-Max} to generate T-CoT annotations for each navigation slice (see Eq.~(\ref{eq:slice})). All 18{,}554 navigation slices from the LH-VLN training set are annotated. The annotation prompt is as follows:
\begin{tcolorbox}[mygraybox]
You are a professional AI data annotator. Your task is to label the intermediate CoT reasoning for VLN trajectory slices. Input includes: user instruction, \(\le\)20 historical images (front view), 3 current images (left, front, and right view), and the GT action sequence. \\

Annotation steps: \\
1.~\textbf{Semantic Planning}: Break the mission into precise sub-tasks with clear spatial goals (e.g., ``reach the cabinet table in the living room,'' ``exit the living room,'' ``enter the office,'' ``approach the office table''). Sub-tasks should reflect stepwise navigation milestones rather than abstract summaries. \\
2.~\textbf{Visual Description}: Based on the semantic plan, describe what the historical and current images reveal about completed and upcoming sub-tasks. Be concise, factual, and avoid vague wording. \\
3.~\textbf{Action Decision-Making}: Predict the next action sequence (5 steps) aligned with the semantic plan and supported by the visual description. Provide a brief justification for the predicted steps. Use natural, instruction-like phrasing instead of raw action codes. \\
4.~\textbf{Visual Imagination}: Describe the expected scene after executing the predicted actions, focusing on landmarks or key objects that should become visible or reachable. \\

\# Output format: \\
\texttt{<think>}\\
Semantic Plan: \(\cdots\)\\
Visual Description: \(\cdots\)\\
Action Planning: \(\cdots\)\\
Visual Imagination: \(\cdots\)\\
\texttt{</think>} \\
Note: Replace ``\(\cdots\)'' with actual CoT content. Do not output quotes or ellipses. \\

\# User Instruction: [Instruction] \\

\# GT Action: \texttt{<|forward|>} \(\cdots\) \texttt{<|right|>}
\end{tcolorbox}

\subsection{Data Augmentation}
To improve the robustness of the instruction-following model under diverse visual histories, we augment each training example by perturbing only the historical image sequence while keeping the final three observation images unchanged. Given an original sample with N historical frames \(\{h_1, \dots, h_N\}\) and the last three observation frames \(\{o_{\ell}, o_{f}, o_{r}\}\), we generate up to two additional augmented variants per sample. The augmentation operations are stochastic and applied with independent Bernoulli trials.

\paragraph{Uniform Subsampling.}
For trajectories with at least ten historical images, we apply uniform subsampling with probability 0.5. Specifically, we replace the original history \(\{h_i\}_{i=1}^{N}\) with a stride-2 subsequence \(\{h_1, h_3, h_5, \dots\}\) while preserving the three observation images. This operation reduces temporal redundancy and encourages the model to rely on coarser but more informative state transitions.

\paragraph{Stochastic History Trimming.}
For trajectories with at least seven historical frames, we further apply stochastic trimming. Two perturbations may occur:
(i) with probability 0.5, we remove the first two historical frames, yielding \(\{h_3, h_4, \dots, h_N\}\);
(ii) with probability 0.5, and only when the remaining length is at least seven, we randomly select an index \(k\) and remove two consecutive frames \(\{h_k, h_{k+1}\}\).
At least one of the above operations must be triggered for the augmented sample to be retained.
This procedure introduces temporal uncertainty, forcing the model to rely on stable, task-relevant cues rather than positional biases.

\section{Detail Formulations.}
Here, we provide the formal formulation of the proposed \textsc{FantasyVLN}, which unifies Non-CoT, T-CoT, V-CoT, and MM-CoT reasoning modes within a single reasoning framework. We introduce binary gating signals \(g_{\mathcal{T}}\) and \(g_{\mathcal{V}}\) to enable flexibly switching among the four reasoning modes.

\paragraph{Non-CoT Reasoning.} Given the task instruction \(\mathcal{I}\) and visual observations \(\{o_{\le t}\}\), the non-CoT reasoning mode aims to directly predict actions \(\widehat{\mathcal{A}}_t\) based on the instruction \(\mathcal{I}\) and observations \(\{o_{\le t}\}\):
\begin{equation}
\widehat{\mathcal{A}}_t \sim \pi_\theta \big( \mathcal{I}, \{o_{\le t}\}, g_\mathcal{T} = 0, g_\mathcal{V} = 0 \big),
\end{equation}
where \(\pi_\theta\) is the navigation agent. We employ a pretrained VLM as the navigation agent \(\pi_\theta\) and transfer its multimodal prior to the navigation task through supervised fine-tuning (SFT):
\begin{equation}
\arg\min_{\theta} \, \mathcal{L}_{\text{CE}}\big(\widehat{\mathcal{A}}_t, \mathcal{A}_t\big),
\end{equation}
where \(\mathcal{L}_{\text{CE}}\) denotes the causal cross-entropy loss and \(\mathcal{A}_t\) represents the ground truth actions.

\paragraph{T-CoT reasoning.} Textual CoT reasoning generates intermediate reasoning steps \(\widehat{\mathcal{T}}_t\) before action prediction:
\begin{equation}
[ \widehat{\mathcal{T}}_t, \widehat{\mathcal{A}}_t ] \sim \pi_\theta \big( \mathcal{I}, \{o_{\le t}\}, g_\mathcal{T} = 1, g_\mathcal{V} = 0 \big).
\end{equation}
We define the intermediate reasoning steps \(\widehat{\mathcal{T}}_t\) as a structured chain of thought that guides the navigation process. The agent \(\pi_\theta\) first decomposes the instruction \(\mathcal{I}\) into a sequence of subgoals, and then infers the current goal from the visual observations \(\{o_{\le t}\}\). 
Finally, \(\pi_\theta\) identifies the decision evidence from the current visual observation \(o_t\).
The training objective is formulated as:
\begin{equation}
\arg \min_\theta \mathcal{L}_{\text{CE}}\big(\widehat{\mathcal{T}}_t, \mathcal{T}_t\big) + \mathcal{L}_{\text{CE}}\big(\widehat{\mathcal{A}}_t, \mathcal{A}_t\big),
\end{equation}
where \(\mathcal{T}_t\) is the ground truth of textual reasoning steps.

\paragraph{CompV-CoT reasoning.} Visual CoT reasoning enhances spatial understanding by generating future visual observations \(\widehat{\mathcal{V}}_t\), which serve as the conditions for action prediction.
We propose CompV-CoT that conducts V-CoT in the latent space of a VAR model. Instead of producing pixel-level images, the agent \(\pi_\theta\) predicts low-scale latent representations of VAR \(\widehat{\mathcal{H}}_t\):
\begin{equation}
[ \widehat{\mathcal{H}}_t, \widehat{\mathcal{A}}_t ] \sim \pi_\theta \big( \mathcal{I}, \{o_{\le t}\}, g_\mathcal{T} = 0, g_\mathcal{V} = 1 \big).
\end{equation}
The predicted representations are then passed through the VAR model to reconstruct pixel observations:
\begin{equation}
\widehat{\mathcal{V}}_t \sim g \big( \widehat{\mathcal{H}}_t \big),
\end{equation}
where \(g\) denotes the generation pipeline based on next-scale prediction for VAR.
During training, the VAR is frozen.
The training process is defined as:
\begin{equation}
\arg \min_\theta \mathcal{L}_{\text{CE}}\big(\widehat{\mathcal{V}}_t, \mathcal{V}_t\big) + \mathcal{L}_{\text{CE}}\big(\widehat{\mathcal{A}}_t, \mathcal{A}_t\big),
\end{equation}
where \(\mathcal{V}_t\) is the ground truth future visual observations.

\paragraph{MM-CoT reasoning}
We employ paired textual-visual CoT reasoning steps as the villain Multimodal CoT reasoning steps \(\widehat{\mathcal{M}}_t\):
\begin{equation}
\widehat{\mathcal{M}}_t = [\widehat{\mathcal{T}}_t, \widehat{\mathcal{H}}_t].
\end{equation}
MM-CoT reasoning first generates multimodal reasoning steps \(\widehat{\mathcal{M}}_t\) and then predicts future actions \(\widehat{\mathcal{A}}_t\):
\begin{equation}
[ \widehat{\mathcal{M}}_t, \widehat{\mathcal{A}}_t ] \sim \pi_\theta \big( \mathcal{I}, \{o_{\le t}\}, g_\mathcal{T} = 1, g_\mathcal{V} = 1 \big).
\end{equation}
The training objective is formulated as:
\begin{equation}
\arg \min_\theta \mathcal{L}_{\text{CE}}\big(\widehat{\mathcal{M}}_t, \mathcal{M}_t\big) + \mathcal{L}_{\text{CE}}\big(\widehat{\mathcal{A}}_t, \mathcal{A}_t\big).
\end{equation}

\paragraph{UM-CoT reasoning.}
\textsc{FantasyVLN} is a unified multimodal CoT reasoning framework that integrates the non-CoT, T-CoT, V-CoT, and MM-CoT reasoning modes. The formulation of UM-CoT is provided in Section~\ref{sec:um-cot}.

\section{Implementation Details}
\textbf{Training Details.}
We adopt \texttt{Qwen2.5-VL} as the base model and apply LoRA-based parameter-efficient tuning to both the language layers and the vision--language projection modules. Training is conducted on 64 H20 GPUs, each equipped with 141\,GB of memory. We use the AdamW optimizer with a learning rate of \(1\times10^{-4}\), a weight decay of \(0.1\), and a cosine schedule with a \(5\%\) warmup ratio. The per-device batch size is set to 4, supported by 32 dataloader workers. We employ \texttt{bfloat16} precision, enable gradient checkpointing, and adopt DeepSpeed ZeRO-2 for memory-efficient optimization.

\textbf{Evaluation.}
We perform online evaluation for all methods. Given an initial position, the agent interacts with the simulator to execute multi-stage navigation tasks. At each step, the agent receives visual observations and predicts subsequent actions; the environment then applies the predicted action and updates the agent’s state. When the agent outputs \texttt{<|stop|>}, the environment verifies whether the current subtask is completed (i.e., the distance to the target location is within 1\,m) and then proceeds to the next subtask. If the agent exhausts its action budget before completing the subtask, the subtask is marked as failed. This procedure continues until all subtasks are completed or terminated.

\paragraph{Special Tokens.}
Leveraging the vocabulary extensibility of autoregressive models, we implement the required functionalities through systematic vocabulary expansion.
Specifically, we introduce (i) action tokens \texttt{<|forward|>}, \texttt{<|left|>}, \texttt{<|right|>}, \texttt{<|stop|>} for navigation action prediction; (ii) VAR latent tokens \texttt{<|1|>}--\texttt{<|4096|>} for CompV-CoT and MM-CoT latent-space visual reasoning; (iii) system tokens such as \texttt{<|NAV|>}, \texttt{<think>}, \texttt{</think>} to regulate narrative formatting; and (iv) gating tokens \texttt{<textual\_think>}, \texttt{<no\_textual\_think>}, \texttt{<visual\_think>}, \texttt{<no\_visual\_think>} that serve as the binary signals \(g_{\mathcal{T}}\) and \(g_{\mathcal{V}}\) for unified multimodal CoT control.

\end{document}